\documentclass[12pt]{article}
\usepackage[utf8]{inputenc}
\usepackage{lineno}

\usepackage{hyperref}       
\usepackage{url}            
\usepackage{amsmath,amsfonts,amsthm}
\usepackage{authblk}
\usepackage{subcaption}
\usepackage{graphicx}
\usepackage{xcolor}
\usepackage{mathrsfs}
\usepackage{mathtools}
\usepackage{booktabs}       %
\usepackage{soul}
\usepackage{xr}
\usepackage[margin=1in]{geometry}
\usepackage{amsthm}
\usepackage{microtype}
\usepackage{graphicx}
\usepackage{capt-of}
\usepackage{subcaption}
\usepackage{booktabs} 
\usepackage{makecell}
\usepackage{amsmath}
\usepackage{authblk}
\usepackage{amssymb}
\usepackage{wrapfig}
\usepackage{fancyvrb}
\usepackage{indentfirst}
\usepackage{multirow}
\usepackage{enumitem}
\usepackage{fancyvrb}
\usepackage{algorithm}

\usepackage[noend]{algpseudocode}
\newcommand{\R}{\mathbb{R}}

\title{Spatio-Temporal Conformal Prediction for \\Power Outage Data}
\author{Hanyang Jiang, Yao Xie, Feng Qiu}
\date{}

\begin{document}

\maketitle

\begin{abstract}
In recent years, increasingly unpredictable and severe global weather patterns have frequently caused long-lasting power outages. Building resilience—the ability to withstand, adapt to, and recover from major disruptions—has become crucial for the power industry. To enable rapid recovery, accurately predicting future outage numbers is essential. Rather than relying on simple point estimates, we analyze extensive quarter-hourly outage data and develop a graph conformal prediction method that delivers accurate prediction regions for outage numbers across the states for a time period. We demonstrate the effectiveness of this method through extensive numerical experiments in several states affected by extreme weather events that led to widespread outages.
\end{abstract}

\section{Introduction}
With the global climate change, extreme weather events like hurricanes, winter storms, and tornadoes have increasingly led to widespread electric power outages across the United States \cite{Handmer2012}. For instance, during March 2018, the northeastern U.S. was battered by three consecutive winter storms within a span of 14 days. This series of events caused power outages that left over 2.75 million customers without electricity in the New England region, resulting in economic losses of approximately \$4 billion, including \$2.9 billion in insured damages \cite{Captive2018}. Such severe weather-related incidents often leave millions without power for extended periods, resulting in significant economic disruption \cite{Executive2013} and, tragically, sometimes even loss of life \cite{Shapiro2017}.

Given the considerable impact of extreme weather on power systems since the early 2000s, regulatory bodies in the U.S. have called on the energy sector to enhance the resilience of power grids through various hardening measures \cite{AbiSamra2013}. Consequently, accurately assessing the resilience of power grids is crucial not only for estimating potential damage from extreme weather but also for informing short-term disaster response strategies, long-term resilience planning, and shaping energy policy.

However, there are significant challenges in studying grid resilience \cite{Jufri2019}. One of the primary obstacles is the disconnect between the data needed by researchers and the data that is actually available. While utility companies and system operators often collect detailed data on power failures and recovery processes, these datasets are typically restricted to internal use and are not widely shared \cite{Bryan2012}. In contrast, U.S. federal and state governments require only aggregated data, such as the total duration of service interruptions, which is often too coarse for detailed resilience studies \cite{Campbell2012}. In addition to data limitations, the lack of sophisticated mathematical models capable of capturing the complex dynamics of power disruptions and restorations under extreme weather conditions further hampers research. The power grid is a highly intricate and interconnected system, where disruptions in one area can rapidly escalate into large-scale blackouts due to the cascading effects across the network \cite{Fairley2004}. As a result, pinpointing the exact factors that lead to widespread outages remains a formidable challenge.

Instead of focusing on the detailed power grid structure, this paper quantitatively estimates power grid resilience under extreme weather by leveraging a unique dataset of fine-grained, customer-level power outage records alongside comprehensive weather data from three major weather events across three U.S. states. Unlike detailed power grid outage records, which are often inaccessible to the public, customer-level outage data reported by local utilities \cite{MEMA2020, Duke2020} are readily available and provide a valuable resource for understanding the dynamics of outages. These records, which track the number of customers without power every 15 minutes across various geographical units (e.g., towns, counties, or zip codes), allow for a detailed analysis of the spatio-temporal interactions between weather-induced disruptions and the subsequent restoration processes. 

With this data, we employed a parametric prediction model to analyze the progression of customer outages without relying on detailed disruption and restoration records. The model captures the complex spatio-temporal dynamics of outages across different geographical units over time using a multivariate random process framework. After training the model on past data, we build a conformal prediction method on top of it to provide an accurate and robust prediction interval for a period in the future. The model captures the internal relationship between different areas. Furthermore, the model introduces a general framework for spatio-temporal uncertainty quantification.

\section{Related Work}
\subsection{Power Grid Resilience}
Research on power grid resilience has garnered significant attention in recent years \cite{Bhusal2020}. Existing studies on resilience evaluation generally fall into two broad categories: qualitative methods and quantitative methods.

Qualitative methods are the most commonly used in resilience evaluation. These approaches typically involve frameworks that provide a high-level overview of resilience at the system or regional level. For example, studies such as \cite{carlson2012resilience} propose frameworks that use surveys, interviews, and rating systems to assess various aspects of resilience, including personal, business, governmental, and infrastructure resilience. The scoring matrix introduced by \cite{Roege2014} offers a method to evaluate system functions from multiple perspectives, and techniques like the analytic hierarchy process (AHP) help translate subjective assessments into quantifiable measures that can aid in decision-making \cite{orencio2013localized}. While these qualitative methods provide a comprehensive overview that can inform long-term energy policies, they are increasingly seen as costly and difficult to implement, particularly given the rise in large-scale blackouts linked to climate change. Additionally, these methods often lack the scientific rigor needed to justify their application in certain scenarios.

Quantitative methods, which focus on the measurable aspects of system performance, are still in the early stages of development, primarily due to the limited availability of real-world data \cite{EPRI2020}. Quantitative approaches to resilience, such as those presented in \cite{Panteli2015modeling}, often involve the introduction of resilience metrics based on detailed disruption and restoration records. These studies generally require granular data on the operational state and infrastructure during various phases of extreme weather events. Alternatively, some researchers have attempted to apply statistical methods in the absence of comprehensive failure and recovery data. For instance, Monte Carlo simulations have been employed to evaluate grid resilience and assess the impact of extreme weather events on the power grid \cite{Baranski2003}. Recent work \cite{dobson2016obtaining} has made progress in examining specific aspects of power grid resilience by utilizing limited aggregated power outage data. \cite{arora2023probabilistic} make a summary of the probabilistic and machine learning methods used in 

\subsection{Conformal Prediction}
Conformal prediction \cite{vovk2005algorithmic} is a widely used framework for distribution-free uncertainty quantification. The process involves defining a "non-conformity score" based on a point estimator $\hat{f}$, such as a prediction residual, which measures how different a potential response $Y$ is from existing data. Scores are calculated on a hold-out set, and the prediction interval includes values of $Y$ whose score is below the $1-\alpha$ quantile of these hold-out scores. This method is popular in regression and classification problems and is distribution- and model-free, requiring no specific assumptions. However, its guarantees rely on exchangeability, which is often violated in time series data. Some recent work \cite{xu2021conformal, xu2024conformal} extend the method to time series cases and show that the conditional coverage is asymptotically valid.

However, there has been very limited research on conformal prediction in spatial context, especially for the spatio-temporal setting. \cite{mao2024valid} recently proposed a spatial conformal prediction method under the infill sampling framework, where data becomes increasingly dense in a bounded region. Their approach relies on assumptions about the spatial process and a locally i.i.d. error structure to establish local exchangeability and derive asymptotic local coverage. Similarly, \cite{guan2023localized} introduced a framework for localized conformal prediction, leveraging a kernel function to measure similarity between data points. However, their method is based on the strong assumption of i.i.d. data, which is rarely encountered in practice. Both methods require the data to be somehow locally i.i.d., which doesn't happen in temporal data. In contrast, our work deals with graph-structured spatio-temporal data, which neither of these approaches can adequately address, as they do not account for the complexities of spatio-temporal dependencies.

\section{Power outage prediction}
\label{sec3}
To predict the number of customer power outages using weather data and historical outage records, we employ a data-driven spatio-temporal model from \cite{zhu2021quantifying}. This model captures the dynamics of outages across a service territory, which is divided into $K$ geographical units over $T$ time slots. We utilize $M$ weather variables to represent the environmental conditions.

\subsection{Model Setup}

Let $i, j \in \{1, \dots, K\}$ denote the indices of geographical units, $t \in \{1, \dots, T\}$ denote the time slots, and $m \in \{1, \dots, M\}$ denote the weather variables. The number of outages in unit $i$ at time $t$ is represented by $N_{it} \in \mathbb{Z}_+$, while $x_{i,t,m} \in \mathbb{R}$ denotes the value of weather variable $m$ in unit $i$ at time $t$.

\subsection{Cumulative Weather Effect}

We define the cumulative weather effect as a weighted sum of recent weather impacts, capturing the decaying influence of past conditions. For weather variable $m$, the cumulative effect $v_{i,t,m}$ in unit $i$ at time $t$ is given by:
\[
v_{i,t,m} = \sum_{\tau=t-d+1}^t x_{i,\tau,m} \exp\{-\omega_m (t - \tau)\},
\]
where $d$ is the time window, $x_{i,\tau,m}$ is the observed value of the weather variable, and $\omega_m \geq 0$ is a learnable decay rate. A larger $\omega_m$ indicates faster decay, whereas $\omega_m = 0$ implies that the weather effect remains constant over time.

\subsection{Outage Occurrence Modeling}

We model the number of outages in each unit as a Poisson process, where $N_{it} \sim \text{Poisson}(\lambda_{it})$. The outage rate $\lambda_{it}$ for unit $i$ at time $t$ depends on both direct weather effects and indirect influences from neighboring units. Let $\boldsymbol{v}_{it} = [v_{i,t,1}, v_{i,t,2}, \dots, v_{i,t,M}]^\top \in \mathbb{R}^M$ denote the vector of accumulated weather effects.

The service area is represented as a directed graph $\mathcal{G} = (\mathcal{V}, \mathcal{E})$, where $\mathcal{V}$ is the set of units and $\mathcal{E}$ is the set of directed edges, indicating the connections between units. The outage rate is defined as:
\[
\lambda_{it} = \gamma_i \mu(\boldsymbol{v}_{it}; \varphi) + \sum_{t' < t} \sum_{(i, j) \in \mathcal{E}} g(i, j, t, t'),
\]
where $\gamma_i \geq 0$ is a scaling factor for unit $i$, and $\mu(\cdot; \varphi) \geq 0$ is a deep neural network-based function, parameterized by $\varphi$, that maps the weather effects to an outage rate. The function $g(i, j, t, t') \geq 0$ models the impact of past outages in neighboring units.

\subsection{Influence Function for Neighboring Units}

The function $g(i, j, t, t')$ quantifies the influence of past outages in unit $j$ at time $t'$ on unit $i$ at time $t$. We adopt a commonly used form where the effect decays exponentially over time \cite{Reinhart2018}:
\[
g(i, j, t, t') = \alpha_{ij} N_{jt'} \beta_j e^{-\beta_j (t-t')}, \quad (i, j) \in \mathcal{E}, \, t > t',
\]
where $\beta_j \geq 0$ is the decay rate, and $\alpha_{ij} \geq 0$ represents the weight of influence from unit $j$ to unit $i$. The parameter $\beta_j$ can also be interpreted as the recovery rate of unit $j$. We set $\alpha_{ii} = 1$ for self-influence, and ensure no loops between distinct units, i.e., $\alpha_{ij} = 0$ if $\alpha_{ji} \neq 0$.

\subsection{Parameter Estimation}

The model parameters, denoted by $\theta = \{\alpha_{ij}\}_{(i, j) \in \mathcal{E}, i \neq j}, \{\beta_i\}_{i \in \mathcal{V}}, \{\gamma_i\}_{i \in \mathcal{V}}, \{\omega_i\}_{i \in \mathcal{V}}, \varphi$, are estimated by maximizing the log-likelihood function:
\[
\ell(\theta | N, X) = - \sum_{t=1}^T \sum_{i=1}^K \left(\lambda_{it}(\theta) - N_{it} \log \lambda_{it}(\theta)\right).
\]
We optimize this objective using stochastic gradient descent, subject to the constraints:
\begin{align*}
\alpha_{ii} &= 1, \quad \forall i \in \{1, \dots, K\}.
\end{align*}
Here, $N$ and $X$ represent the observed outage counts and weather data, respectively.

\section{Uncertainty Quantification on Time Series Data with Graph Structure}
\subsection{Problem setup}
Given a weighted graph $G=(V,E)$, where $|V|=K$ is the number of geographical units in a specific service area and each edge in $E$ is assigned a specific weight. Assume a sequence of observations $(X_t,Y_t), t=1,2,\cdots$ where $Y_t\in \R^{K}$, $X_t\in\R^{K\times M}$ are continuous vectors. The i-th row of $X_t$ is the feature vector $X_{it}\in\R^M$ of unit $i$, and $Y_{it}$ is the response at unit $i$. In our problem, the feature vector $X_{it}$ can be the location, weather condition or the history of $Y_{it}$. The response $Y_{it}$ records the number of power outage at unit $i$.

\subsection{Vanilla Conformal Prediction}
Suppose we have a dataset of observations $\{(X_t, Y_t)\}_{t=1}^{T}$, where $X_t$ represents the features and $Y_t$ the response variable at time $t$. Our goal is to construct a prediction interval $\hat{C}_{T}(X_{T+1})$ for the future observation $Y_{T+1}$ such that it contains $Y_{T+1}$ with a target coverage probability of $1-\alpha$. Here, $\alpha \in (0,1)$ is the user-specified significance level that controls the allowable error rate, though we will omit $\alpha$ in the notation for simplicity.

Given a pre-trained prediction model $\hat{f}$, which in our case is the Poisson model introduced in Section \ref{sec3}, we can compute the prediction residuals, also known as non-conformity scores, for each observation:
\begin{equation}
\hat{\varepsilon}_t = |Y_t - \hat{f}(X_t)|, \quad t = 1, \ldots, T.
\end{equation}
These non-conformity scores measure how different the observed response $Y_t$ is from the predicted value $\hat{f}(X_t)$, providing a way to assess the prediction uncertainty.

Conformal prediction aims to satisfy the following \textit{marginal coverage} property under the assumption of exchangeability:
\begin{equation}
\mathbb{P}(Y_{T+1} \in \hat{C}_{T}(X_{T+1})) \ge 1-\alpha.
\end{equation}
If this inequality holds, then the prediction interval $\hat{C}_{T}(X_{T+1})$ is considered to be \textit{marginally valid}, meaning that, on average, the interval will cover the true value of $Y_{T+1}$ with at least probability $1-\alpha$ across all possible realizations of the data.

The \textit{exchangeability} assumption implies that the joint distribution of the data remains unchanged under any permutation:
\begin{equation}
f((X_1, Y_1), \ldots, (X_s, Y_s)) = f((X_{\sigma(1)}, Y_{\sigma(1)}), \ldots, (X_{\sigma(s)}, Y_{\sigma(s)})),
\end{equation}
for any integer $s$ and any permutation $\sigma$ of the indices. This assumption is weaker than the assumption of independence and identical distribution (i.i.d.) and allows for the validity of conformal prediction methods in a broader range of settings.

The goal in conformal prediction is not only to ensure marginal validity but also to construct a prediction interval that is as narrow as possible while satisfying the coverage requirement. It is trivial to achieve marginal validity by setting the prediction region to the entire outcome space, but such an interval would be uninformative.

A commonly used approach, known as \textit{full conformal prediction}, constructs the prediction interval based on the empirical distribution of the non-conformity scores. Specifically, the prediction interval for a new observation $Y_{T+1}$ is given by:
\begin{equation}
\hat{C}_{T}(X_{T+1}) = \left\{ y : \hat{\varepsilon}_{T+1} \leq Q_{1-\alpha}\left( \sum_{t=1}^{T+1} \frac{1}{T+1} \delta_{\hat{\varepsilon}_t} \right) \right\},
\end{equation}
where $Q_{1-\alpha}(\cdot)$ denotes the $1-\alpha$ quantile of the empirical distribution, and $\delta_{\varepsilon_t}$ is the Dirac measure at $\varepsilon_t$. This formulation treats all observed data points and the test data equally when determining the threshold for inclusion in the prediction interval.

The validity of this approach is guaranteed under the exchangeability assumption, meaning that split conformal prediction achieves the desired coverage probability across different data samples. However, when the data exhibit dependencies (e.g., time series or spatial data), the exchangeability assumption may be violated, which poses a challenge for applying traditional conformal prediction methods. In such cases, modifications are needed to account for the dependence structure in the data.

\subsection{Spatio-temporal Conformal prediction for graph structure}
In the context of power outage prediction, observations are often highly correlated in both time and space, violating the independence or exchangeability assumptions required by standard conformal prediction methods. Although there are conformal prediction techniques developed specifically for time series data \cite{xu2023sequential}, they typically do not account for the spatial relationships between different locations. To address these dependencies, existing extensions of conformal prediction incorporate weighted schemes or modify quantile estimation to account for the correlation structure in the data, aiming to maintain valid prediction intervals with minimal width.

To better leverage the graph structure of the problem, we propose a Graph Conformal Prediction (Graph CP) method, outlined in Algorithm \ref{alg}, which provides a more flexible framework for handling spatio-temporal data. Our approach explicitly considers the temporal dependencies of the residuals as well as the spatial dependencies among connected nodes when constructing prediction intervals.

Let the non-conformity score be defined as $\hat{\varepsilon}_{it} = |Y_{it} - \hat{f}(X_{it})|$, where $i$ indexes the nodes in the graph and $t$ represents the time index. Given a window size $\omega$, we define the history of non-conformity scores for node $i$ as $\mathcal{E}_{it}^\omega = \{\hat{\varepsilon}_{i(t-1)}, \ldots, \hat{\varepsilon}_{i(t-\omega)}\}$. The conditional cumulative distribution function (CDF) of the non-conformity score given its history is denoted by $F(z | \mathcal{E}_{it}^\omega) = \mathbb{P}(\hat{\varepsilon}_{it} \leq z | \mathcal{E}_{it}^\omega)$. Due to the correlation between data points, this conditional distribution may differ from the marginal distribution of the residuals. The conditional quantile at level $p$ is then defined as:
\begin{equation}
    Q_t(p) = \inf \{z : F(z | \mathcal{E}_{it}^\omega) \geq p \}.
\end{equation}

Let $\hat{Q}_T(p)$ be an estimate of the true conditional quantile $Q_T(p)$. The Graph CP prediction interval for the future observation is given by:
\begin{equation}
    \hat{C}_{T+1}(X_{T+1}) = \left[\hat{f}(X_{T+1}) + \hat{Q}_T(\alpha/2), \, \hat{f}(X_{T+1}) + \hat{Q}_T(1-\alpha/2)\right].
\end{equation}

A crucial step in the framework is to accurately estimate the true conditional quantile. In classical quantile regression, the quantile function $\hat{Q}_T$ is estimated by minimizing the pinball loss:
\begin{equation}\label{quantile_loss}
    \mathcal{L}(x, \alpha) = 
    \begin{cases}
        \alpha x, & \text{if } x \geq 0, \\
        (\alpha - 1) x, & \text{if } x < 0,
    \end{cases}
\end{equation}
where $\alpha$ is the significance level. However, repeatedly fitting quantile regression models for each time step can be computationally expensive. To improve efficiency, we use Quantile Random Forest (QRF) \cite{meinshausen2006quantile}, which allows for rapid estimation of quantiles.

The QRF method effectively assigns different weights to historical data, thereby leveraging temporal dependencies. Specifically, we use the past $\omega \geq 1$ residuals to predict the conditional quantile of the future residual. To further incorporate spatial dependencies, we assume that the distribution of residuals is similar across neighboring nodes. This assumption motivates us to train the quantile regression model on the set of nodes directly connected to node $i$, denoted as $N(i)$.

Define $\tilde{T} := T - \omega$. For each time step $t' = 1, \ldots, \tilde{T}$, let the input features and target for quantile regression be:
\begin{equation}\label{QRF_data}
    \tilde{X}_{it'} := [\hat{\varepsilon}_{i(t' + \omega - 1)}, \ldots, \hat{\varepsilon}_{it'}], \quad \tilde{Y}_{it'} := \hat{\varepsilon}_{i(t' + \omega)}.
\end{equation}
The feature vector $\tilde{X}_{it'}$ contains $\omega$ past residuals, which are used to predict the quantile of the future residual $\tilde{Y}_{it'}$ at time $t' + \omega$. To estimate the conditional quantile of $\tilde{Y}_{i(\tilde{T} + 1)}$, we use the feature vector $\tilde{X}_{i(\tilde{T} + 1)}$ as input. To leverage the similarity between connected nodes, the QRF is trained using $|N(i)| \tilde{T}$ training data points, $(\tilde{X}_{it'}, \tilde{Y}_{it'})$, where $t' = 1, \ldots, \tilde{T}$ and $i \in N(i)$. At each prediction time step, we update the training data using a sliding window of the most recent $T$ residuals. In our implementation, we utilize the Python QRF package provided by \cite{sklearn_quantile} for efficient computation.

\begin{algorithm}[t]
\caption{Graph Conformal Prediction}
\label{alg}
\begin{algorithmic}
\Require Calibration data $\left\{\left(x_{it}, y_{it}\right)\right\} (1\le i\le K, 1\le t\le T)$, pre-trained prediction algorithm $\hat{f}$, significance level $\alpha$, window size $\omega$
\Ensure Prediction intervals $\{\hat{C}_T\left(x_{j(T+1)}\right)\}$
\For{$1\le j\le K$}:
    \State Select the neighbors of node $j$ as $N(j)=\{n_1,\cdots,n_{j_m}\}$
    \State Compute the prediction residuals $\hat{\varepsilon}=y-\hat{f}(x)$ on the calibration data $\{(x_{it},y_{it})\}_{i\in N(j),t\le T}$
    \State Train quantile regression to obtain $\hat{Q}_t$
    \State Compute the prediction residuals $\widehat{C}_{T'-1}(X_{jT'})$
    \State Obtain new residual $\hat{\varepsilon}_{jT'}$
    \State Update residual history $\mathcal{E}_{iT'}^{\omega}$
\EndFor
\end{algorithmic}
\end{algorithm}

\section{Numerical Experiments}
\subsection{Setting}
In our experiment, we aim to predict power outages during extreme weather events using a unique dataset that combines detailed customer-level outage records with comprehensive weather data. The dataset covers three major service regions spanning four U.S. East Coast states—Massachusetts, Georgia, North Carolina, and South Carolina—over an area of more than 155,000 square miles. The analysis focuses on three significant extreme weather events: the March 2018 nor'easters in Massachusetts, Hurricane Michael in Georgia in October 2018, and Hurricane Isaias in North and South Carolina in August 2020.

\paragraph{March 2018 Nor'easters} 
In March 2018, a series of three powerful winter storms, known as the nor'easters, struck the northeastern, mid-Atlantic, and southeastern regions of the United States, causing extensive damage. At the height of the storms, over $14\%$ of Massachusetts customers (385,744) were without power, and more than $50\%$ of the geographic units (104) experienced significant outage rates.

\paragraph{Hurricane Michael}
Hurricane Michael made landfall in October 2018 as the first Category 5 hurricane to hit the contiguous U.S. since Hurricane Andrew in 1992, resulting in widespread devastation. In Georgia, approximately $7.5\%$ of customers (193,018) experienced power outages, with more than $50\%$ outage rates reported in about $25\%$ of the geographic units (166).

\paragraph{Hurricane Isaias}
In August 2020, Hurricane Isaias, a Category 1 hurricane, caused extensive damage along the Caribbean and U.S. east coast, accompanied by a significant tornado outbreak. At its peak, around $13\%$ of customers in North and South Carolina lost power, with outage rates exceeding $50\%$ in nearly $9\%$ of the geographic units.

\paragraph{Data Splitting and Model Training}
The dataset is divided into three parts: a training set, a calibration set, and a test set, each comprising approximately one-third of the data. The training and calibration sets are selected to include dates when extreme weather events occurred, ensuring that the model learns the system's behavior under severe conditions. A Poisson regression model is first trained on the training set to learn the relationship between weather variables and power outages. The trained model is then used to predict outage occurrences on the calibration and test sets, utilizing both weather data and outage history.

\begin{figure}
    \centering
    \includegraphics[width=0.3\linewidth]{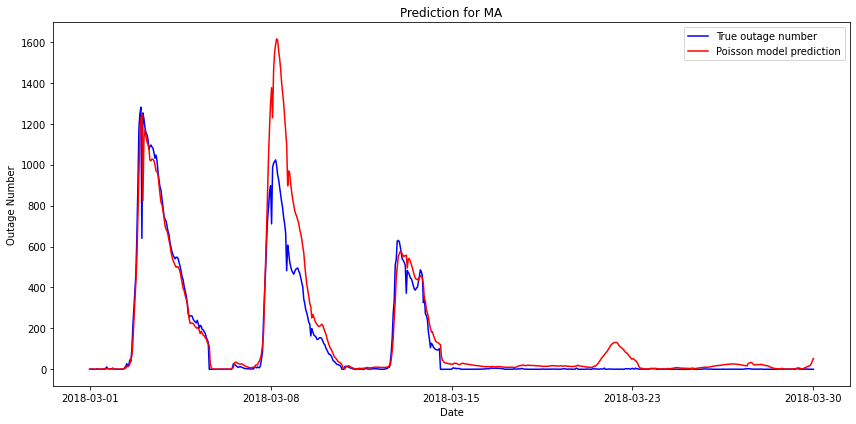}
    \includegraphics[width=0.3\linewidth]{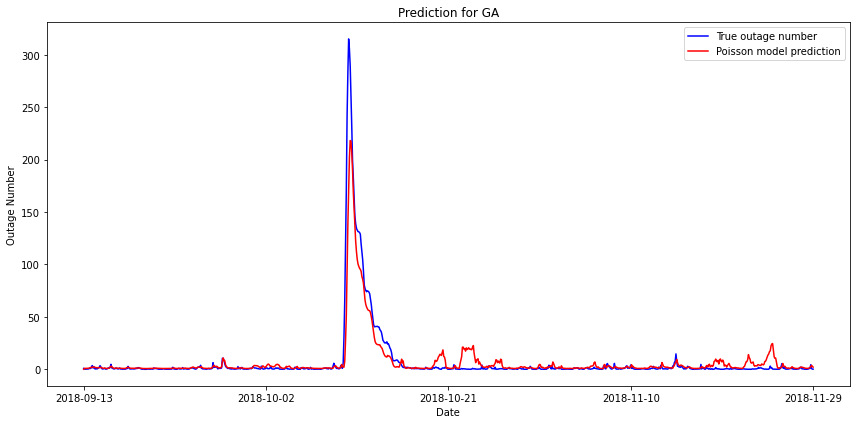}
    \includegraphics[width=0.3\linewidth]{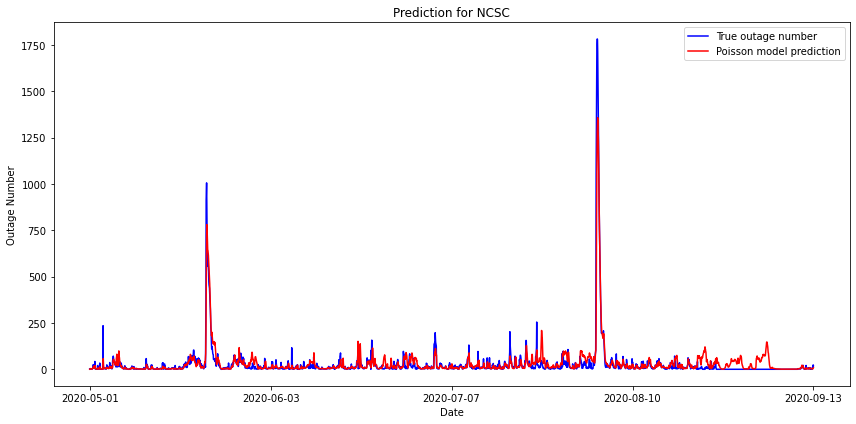}
    \caption{The Possion model prediction for three states.}
    \label{pred}
\end{figure}

\begin{figure}
    \centering
    \includegraphics[width=0.22\linewidth]{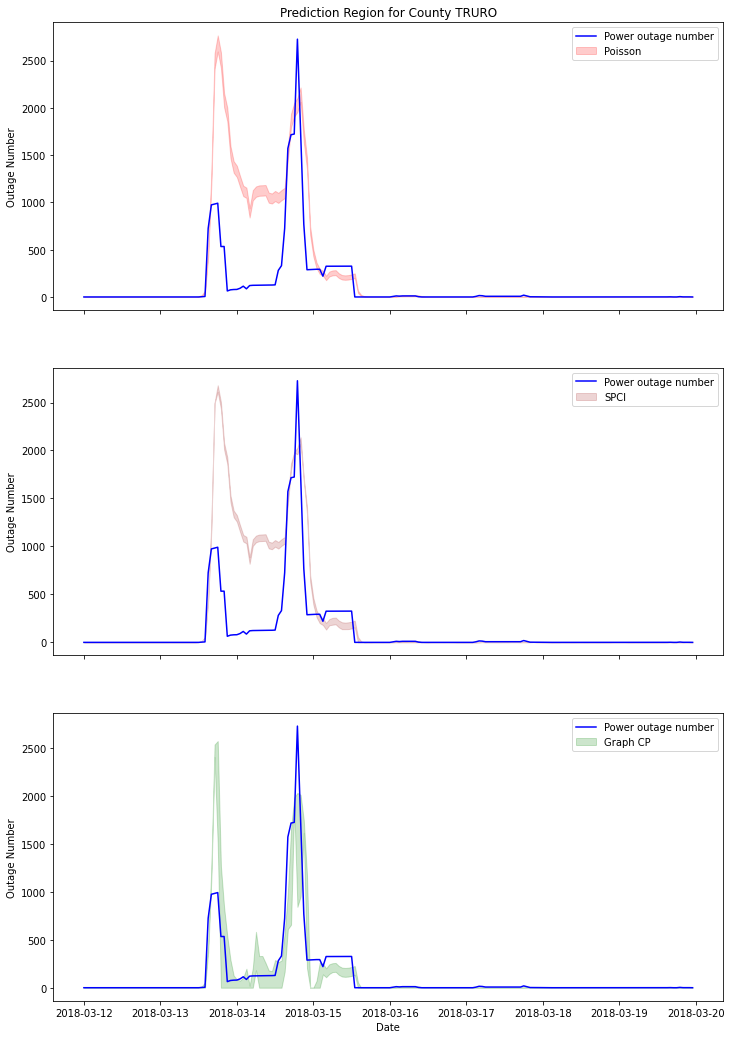}
    \includegraphics[width=0.22\linewidth]{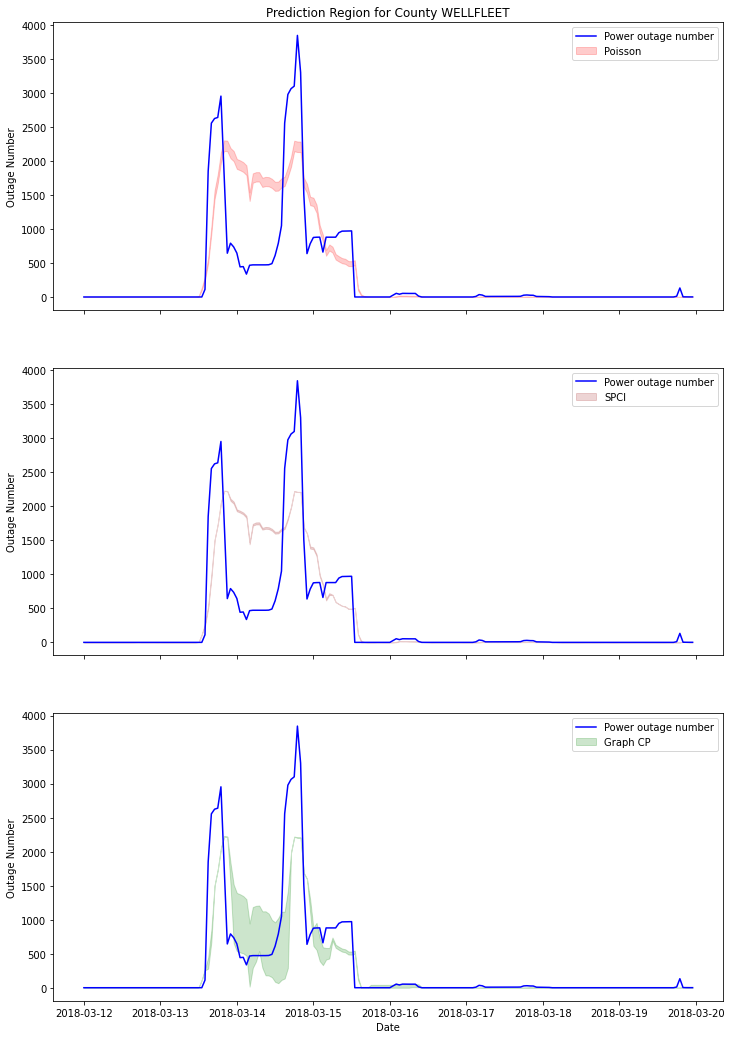}
    \includegraphics[width=0.22\linewidth]{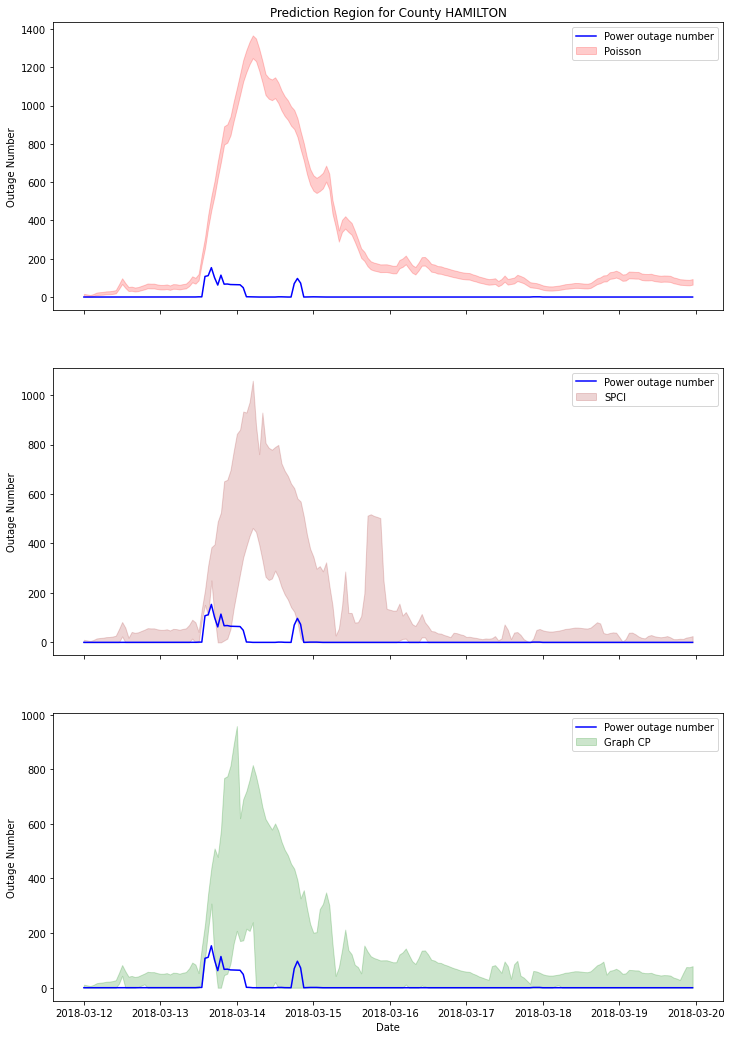}
    \includegraphics[width=0.22\linewidth]{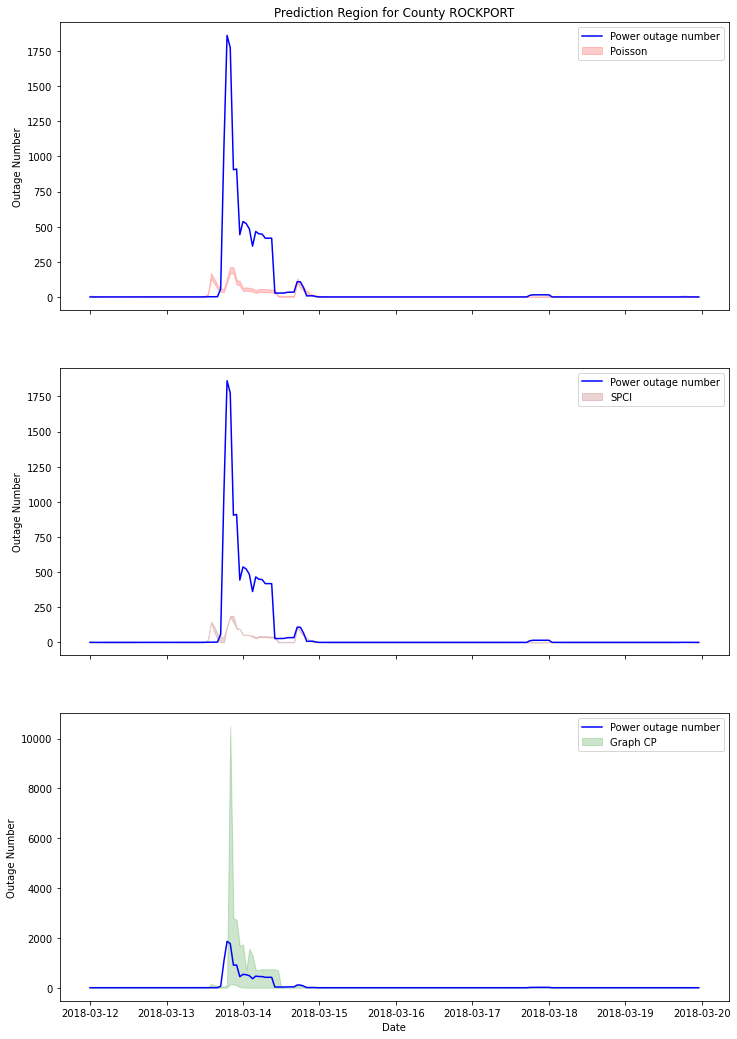}
    \caption{The $90\%$ prediction regions for different counties in Massachusetts.}
    \label{ma_pr}
\end{figure}

\subsection{Real-data results}
\paragraph{Predictions Across the Three States}
Figure \ref{pred} illustrates the Poisson model's predictions for power outages during extreme weather events across three states. The model was trained using a dataset that includes both days of extreme weather and normal conditions. The results indicate that the model is able to capture significant power outages based on weather data and historical outage patterns. The non-zero outage coverage shown in the tables represents the percentage of days where power outages exceeded zero.

\begin{figure}
    \centering
    \begin{minipage}{0.3\linewidth}
        \centering
        \includegraphics[width=\linewidth]{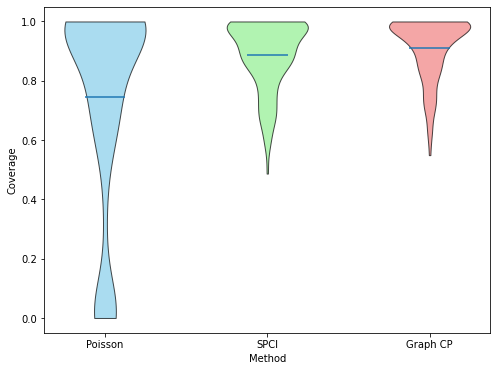}
        \subcaption{Massachusetts}
    \end{minipage}
    \begin{minipage}{0.3\linewidth}
        \centering
        \includegraphics[width=\linewidth]{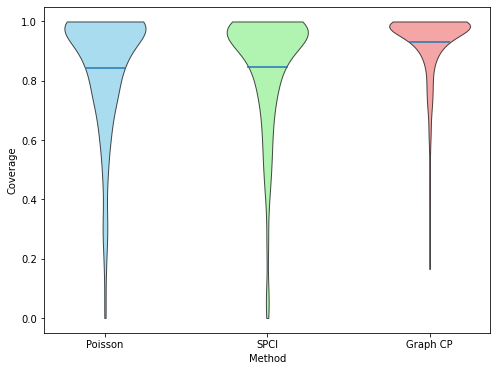}
        \subcaption{Georgia}
    \end{minipage}
    \begin{minipage}{0.3\linewidth}
        \centering
        \includegraphics[width=\linewidth]{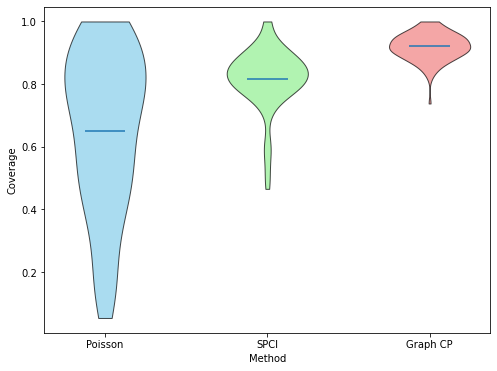}
        \subcaption{North and South Carolina}
    \end{minipage}
    \caption{The violin plots illustrate the coverage distribution for the three methods across different areas. A more concentrated distribution indicates that the method provides more consistent predictions across various regions.}
    \label{violin}
\end{figure}

\paragraph{Comparison with the Poisson Model}
The Poisson model, being a parametric approach, does not offer guaranteed coverage. As seen in Table \ref{combined_table}, the model consistently underestimates the extent of outages, especially when predicting non-zero outages. This issue arises because the model frequently predicts values near zero. While it can successfully predict scenarios with no outages, it struggles to cover actual outage events. In contrast, the two conformal prediction methods provide significantly better overall coverage, including for non-zero outages. Although the Poisson model produces narrower prediction intervals, these intervals are of limited value due to their inadequate coverage. The violin plot in Figure \ref{violin} further demonstrates the model's inconsistency, with coverage levels varying widely from 0 to 1 across different regions. Moreover, Table \ref{wr}, which compares the methods in terms of their best performance across regions, shows that the Poisson model consistently underperforms compared to the other approaches.

\begin{figure}
    \centering
    \includegraphics[width=0.22\linewidth]{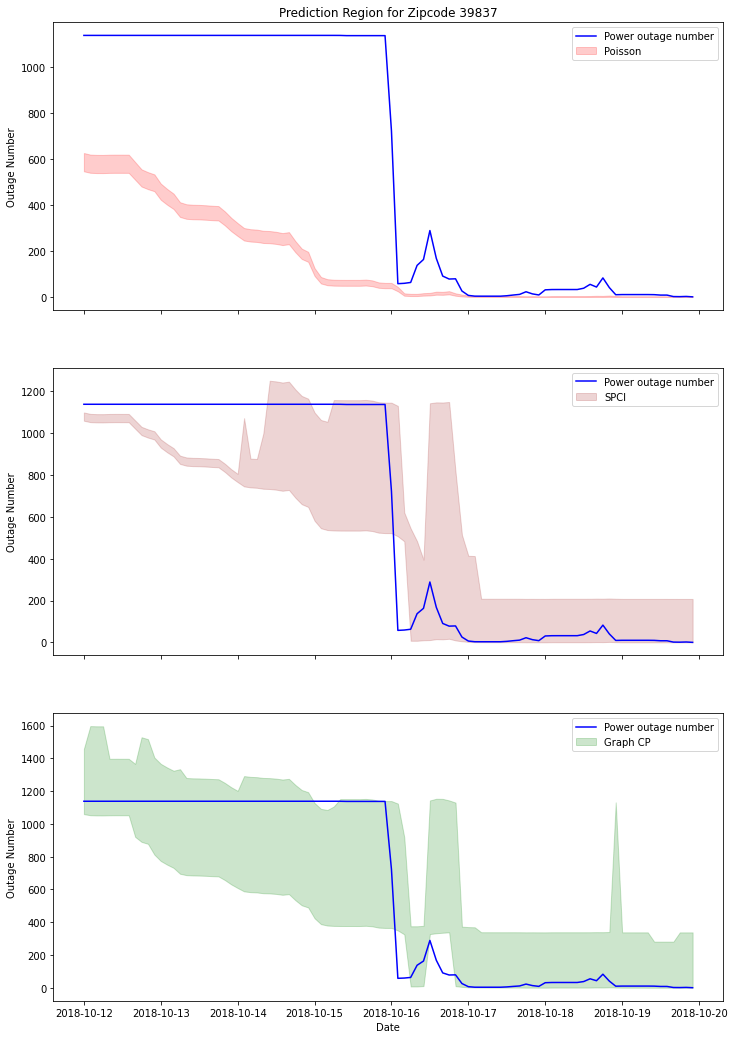}
    \includegraphics[width=0.22\linewidth]{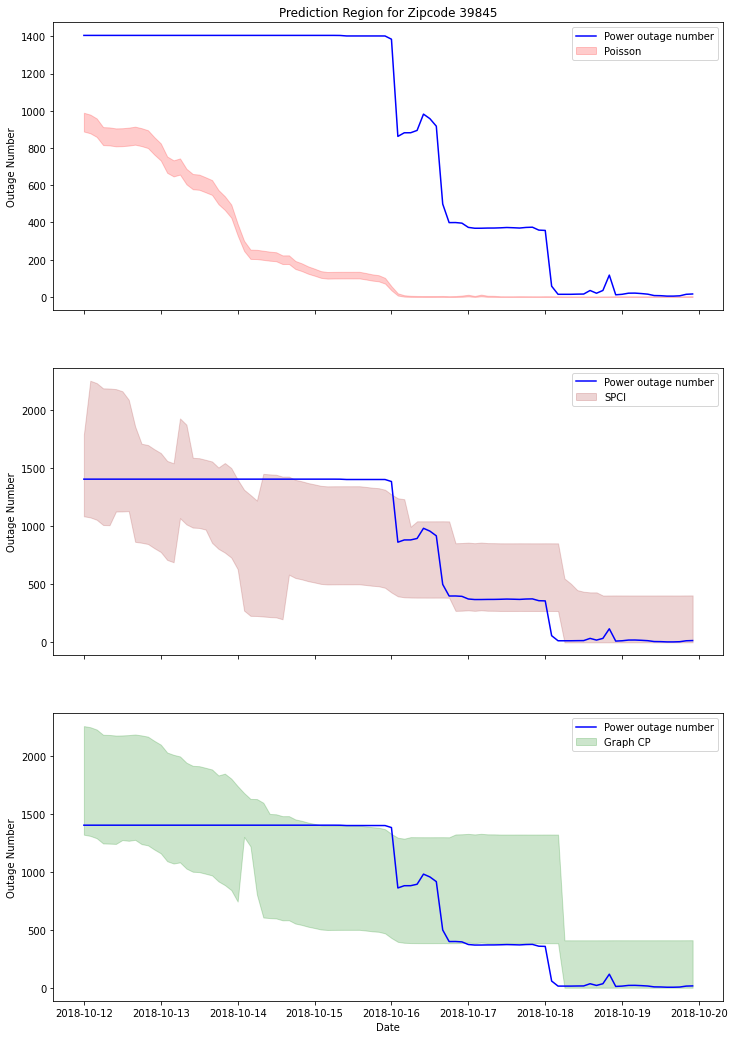}
    \includegraphics[width=0.22\linewidth]{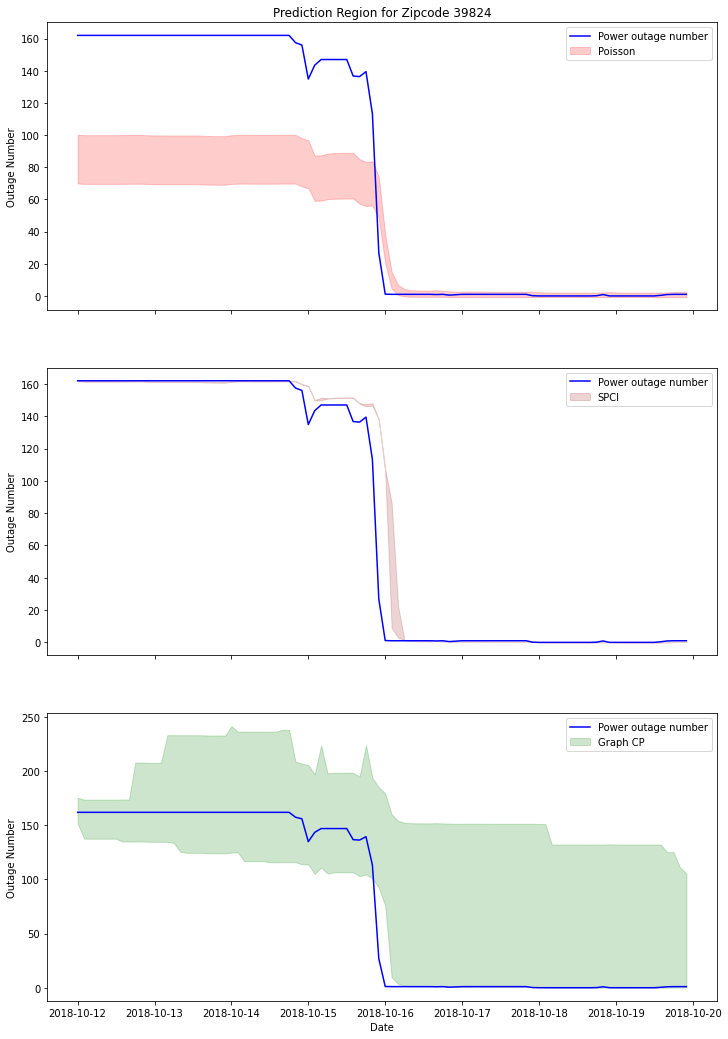}
    \includegraphics[width=0.22\linewidth]{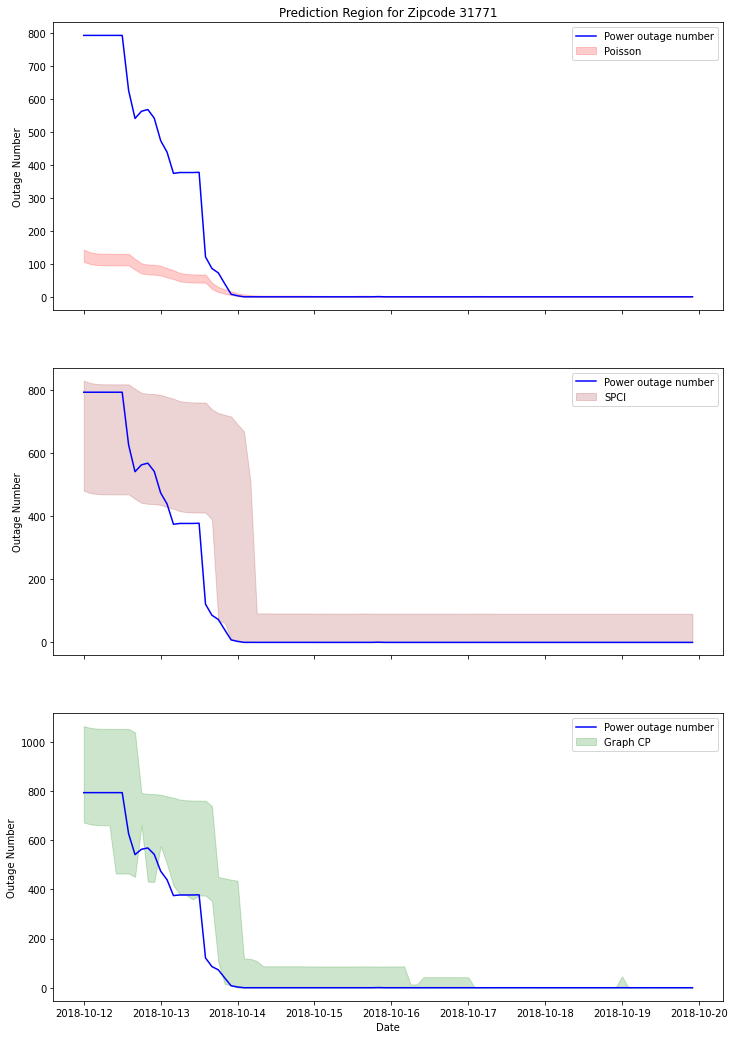}
    \caption{The $90\%$ prediction regions for different zip code areas in Georgia.}
    \label{ga_pr}
\end{figure}

\begin{figure}
    \centering
    \includegraphics[width=0.22\linewidth]{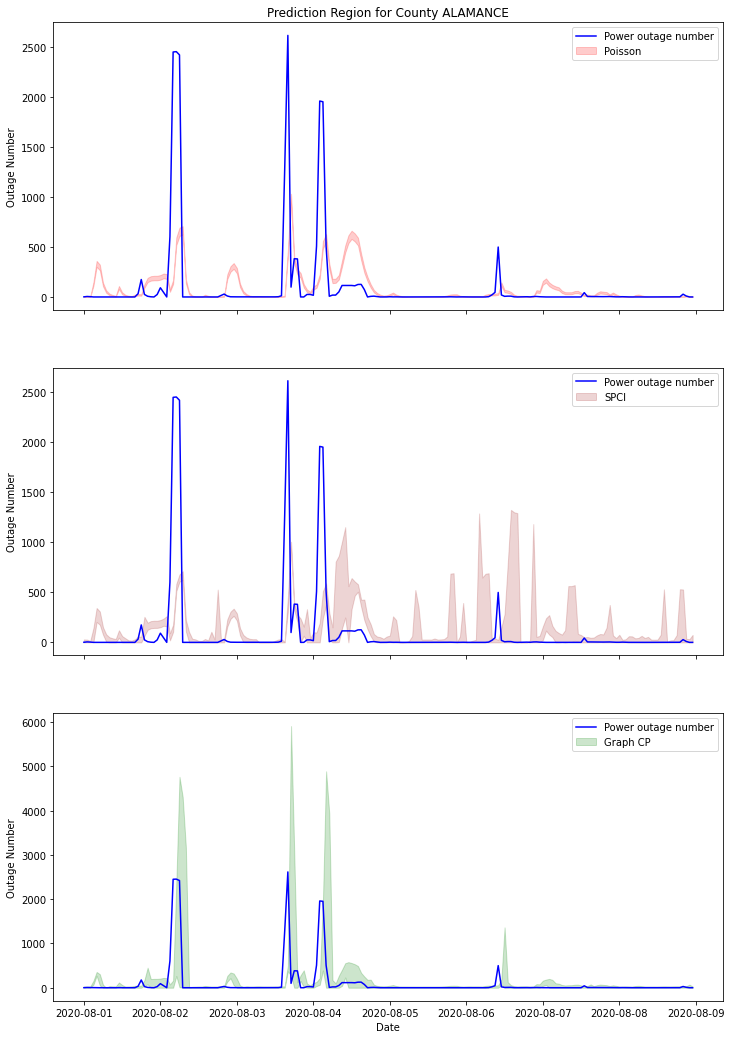}
    \includegraphics[width=0.22\linewidth]{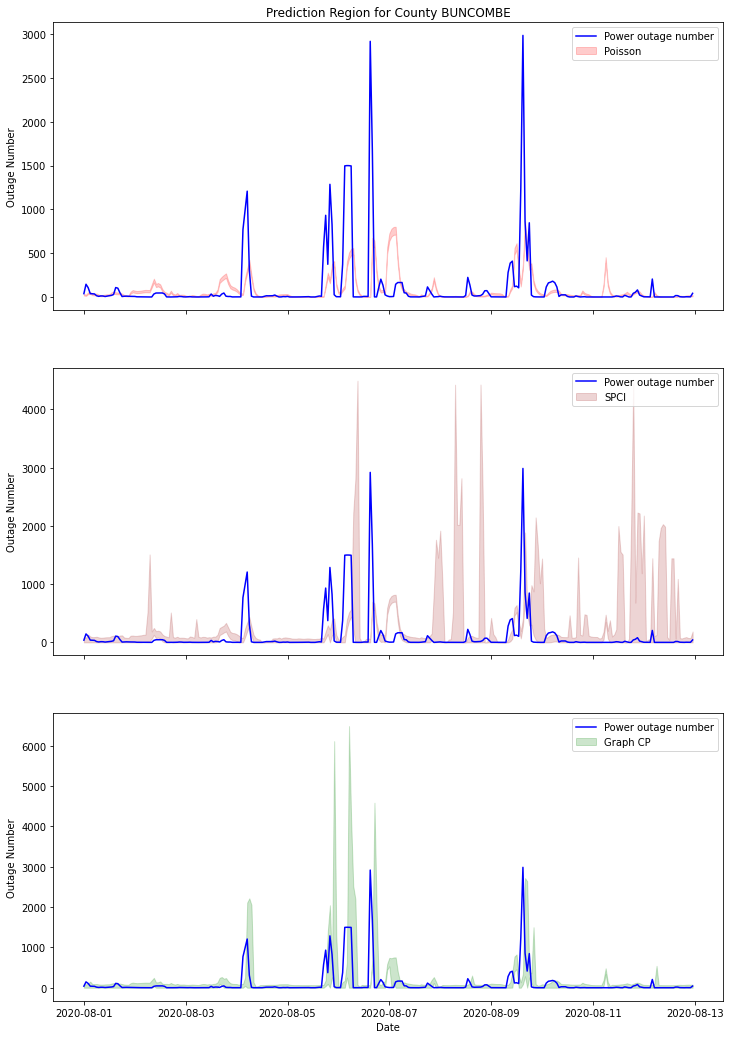}
    \includegraphics[width=0.22\linewidth]{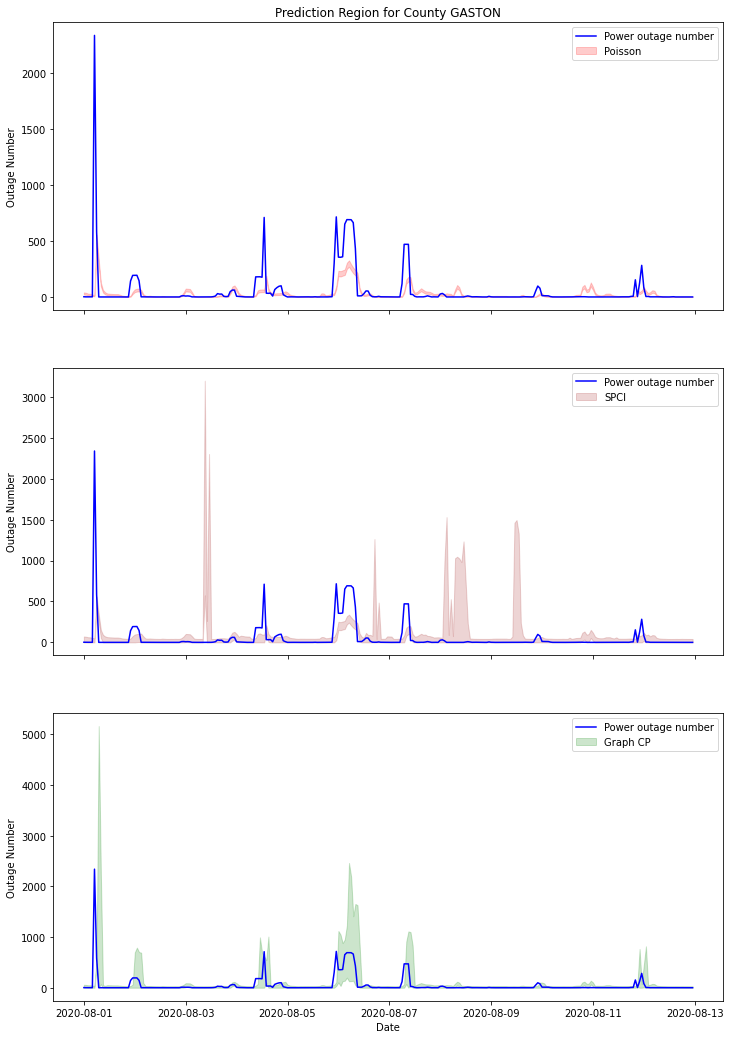}
    \includegraphics[width=0.22\linewidth]{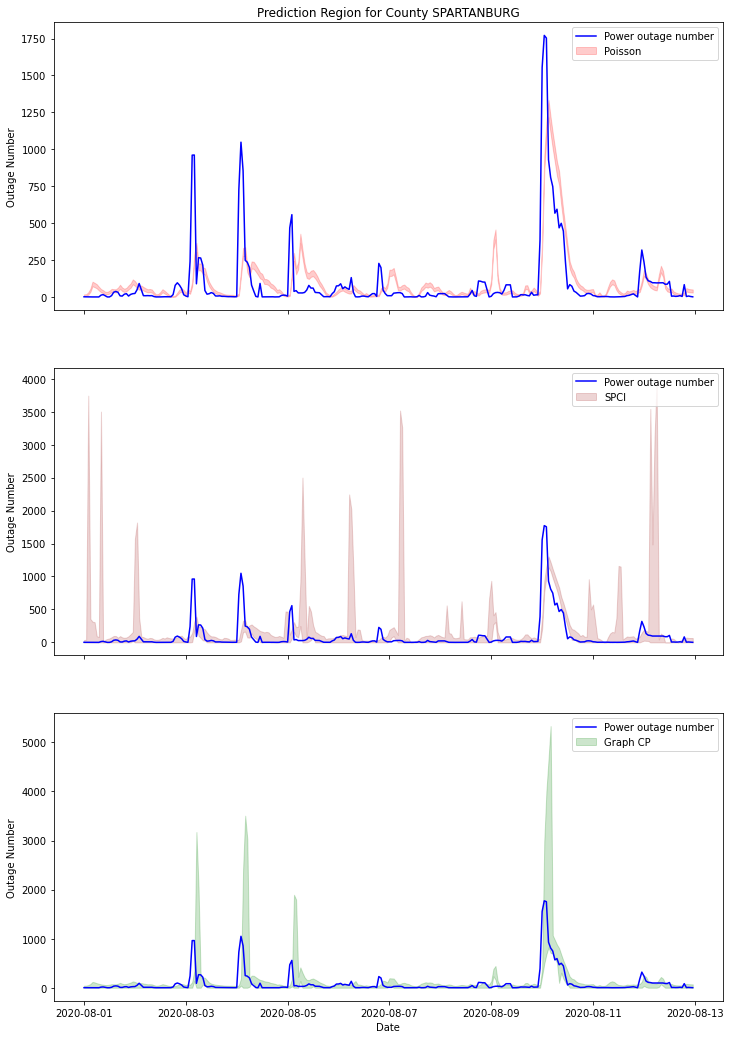}
    \caption{The $90\%$ prediction regions for different zip code areas in North and South Carolina.}
    \label{nc_pr}
\end{figure}

\paragraph{Comparison with SPCI}
SPCI is a time series conformal prediction method that has good empirical performance. However, it still tends to undercover the true power outages, largely because it only accounts for temporal dependencies. Extreme weather events are rare in the data and often cause abrupt distribution shifts that SPCI struggles to capture accurately in a short time frame with limited data. In contrast, our proposed Graph CP method achieves the target coverage in all experiments. By leveraging the spatial relationships among different areas, Graph CP incorporates additional information that enables it to better adapt to distributional changes. The violin plot indicates that Graph CP also provides more consistent coverage than SPCI. Although the prediction intervals from Graph CP are slightly wider, they come with substantially improved coverage.

\begin{table*}[!t]
    \caption{The table presents the proportion of times each method performs the best among the three approaches. Only regions with an average outage number exceeding 50 are considered. A method is deemed the winner if it is the only one to achieve the target coverage in a given region. When multiple methods meet the target coverage, the method with the narrowest interval is selected as the winner. If none of the methods achieve the target coverage, the method with the highest coverage is chosen.}
    \label{wr}
    \centering
    \begin{tabular}{lccc}
    \toprule
    Method & Massachusetts & Georgia & North and South Carolina \\
    \midrule
    Poisson & 6\% & 0\% & 0\% \\
    SPCI & 32\% & 25\% & 9\%\\
    Graph CP & \textbf{62\%} & \textbf{75\%} & \textbf{91\%} \\
    \bottomrule
    \end{tabular}
    \vspace{-0.05in}
\end{table*}

\begin{table*}[!t]
    \caption{Comparison of different methods across scenarios (MA, GA, and NCSC). The table shows coverage, coverage of non-zero outages, and interval width for various methods.}
    \label{combined_table}
    \centering
    \begin{tabular}{llccc}
    \toprule
    State & Method & Coverage & Coverage of Non-Zero Outage & Width \\
    \midrule
    MA & Poisson & 74.5\% & 3.3\% & 12.7 \\
       & SPCI & 88.7\% & 6.8\% & 24.9 \\
       & Graph CP & \textbf{91.1\%} & \textbf{8.1\%} & 37.2 \\
    \midrule
    GA & Poisson & 84.4\% & 3.5\% & 6.15 \\
       & SPCI & 84.5\% & 8\% & 170.5 \\
       & Graph CP & \textbf{93.1\%} & \textbf{10.4\%} & 182.8 \\
    \midrule
    NCSC & Poisson & 64.9\% & 6.3\% & 9.98 \\
         & SPCI & 81.8\% & 11.4\% & 73.5 \\
         & Graph CP & \textbf{92.1\% }& \textbf{15.1\%} & 89 \\
    \bottomrule
    \end{tabular}
    \vspace{-0.05in}
\end{table*}

\section{Discussion}

The results of our study demonstrate the effectiveness of the Graph Conformal Prediction (Graph CP) method for predicting power outages during extreme weather events. The proposed approach addresses several limitations of existing parametric models and conformal prediction techniques, especially in scenarios where spatial and temporal dependencies are significant.

One of the main challenges in power outage prediction is the presence of strong dependencies in both time and space. Standard conformal prediction methods typically assume independent and identically distributed (i.i.d.) data, or at least exchangeability, which does not hold for power outage data due to correlated weather events and interconnected grid dynamics. Existing time-series conformal prediction methods such as SPCI account for temporal correlations but fail to capture the spatial relationships between different regions. 

Graph CP overcomes these limitations by explicitly modeling the spatio-temporal dependencies using a graph structure that represents the connectivity between geographical units. This framework not only accounts for local dependencies but also allows for leveraging information from neighboring areas, thereby providing more accurate and consistent prediction intervals. Our results indicate that Graph CP achieves target coverage in all tested cases, outperforming both the Poisson model and SPCI, particularly in regions with higher outage rates.

The superior performance of Graph CP in predicting power outages has significant implications for power grid resilience planning and emergency response. Accurate prediction intervals provide utility companies with reliable estimates of outage risks, enabling them to allocate resources more effectively for preventive measures and post-event restoration. Moreover, the ability to quantify prediction uncertainty is crucial for decision-makers who need to balance risk and operational costs in preparing for extreme weather events.

While Graph CP shows promise, there are areas for improvement. The current approach relies on a fixed graph structure, which may not fully capture the dynamic nature of power grid interactions during extreme events. Our framework can be easily extended to cases where the graph has weighted edges, measuring the similarity across areas.

\newpage
\bibliographystyle{plain}
\bibliography{reference}

\begin{thebibliography}{10}

\bibitem{AbiSamra2013}
Nicholas Abi-Samra, Lee Willis, and Marvin Moon.
\newblock {\em Hardening the System}, 2013.

\bibitem{MEMA2020}
Massachusetts Emergency~Management Agency.
\newblock {\em Massachusetts Power Outages}, 2020.

\bibitem{arora2023probabilistic}
Prateek Arora and Luis Ceferino.
\newblock Probabilistic and machine learning methods for uncertainty quantification in power outage prediction due to extreme events.
\newblock {\em Natural Hazards and Earth System Sciences}, 23(5):1665--1683, 2023.

\bibitem{Baranski2003}
Michael Baranski and J{\"u}rgen Voss.
\newblock Nonintrusive appliance load monitoring based on an optical sensor.
\newblock In {\em 2003 IEEE Bologna Power Tech Conference Proceedings,}, volume~4, pages 8--pp. IEEE, 2003.

\bibitem{Bhusal2020}
Narayan Bhusal, Michael Abdelmalak, MD~Kamruzzaman, and Mohammed Benidris.
\newblock Power system resilience: Current practices, challenges, and future directions.
\newblock {\em IEEE Access}, 8:18064--18086, 2020.

\bibitem{Bryan2012}
William~N. Bryan.
\newblock {\em Hurricane Sandy Situation Report 20 (U.S. Department of Energy Office of Electricity Delivery \& Energy Reliability)}, 2012.

\bibitem{Campbell2012}
Richard~J Campbell and Sean Lowry.
\newblock Weather-related power outages and electric system resiliency, 2012.

\bibitem{Captive2018}
Captive.com.
\newblock {\em Global Economic Losses \$36 Billion So Far in 2018, over Half Insured}, 2018.

\bibitem{carlson2012resilience}
JL~Carlson, RA~Haffenden, GW~Bassett, WA~Buehring, MJ~Collins~III, SM~Folga, FD~Petit, JA~Phillips, DR~Verner, and RG~Whitfield.
\newblock Resilience: Theory and application.
\newblock Technical report, Argonne National Laboratory, Lemont, IL (United States), 2012.

\bibitem{dobson2016obtaining}
Ian Dobson, Benjamin~A Carreras, David~E Newman, and Jos{\'e}~M Reynolds-Barredo.
\newblock Obtaining statistics of cascading line outages spreading in an electric transmission network from standard utility data.
\newblock {\em IEEE Transactions on Power Systems}, 31(6):4831--4841, 2016.

\bibitem{Duke2020}
Duke Energy.
\newblock {\em Outage Map}, 2020.

\bibitem{Fairley2004}
Peter Fairley.
\newblock The unruly power grid.
\newblock {\em IEEE Spectrum}, 41(8):22--27, 2004.

\bibitem{guan2023localized}
Leying Guan.
\newblock Localized conformal prediction: A generalized inference framework for conformal prediction.
\newblock {\em Biometrika}, 110(1):33--50, 2023.

\bibitem{Handmer2012}
John Handmer, Yasushi Honda, Zbigniew~W Kundzewicz, Nigel Arnell, Gerardo Benito, Jerry Hatfield, Ismail~Fadl Mohamed, Pascal Peduzzi, Shaohong Wu, Boris Sherstyukov, et~al.
\newblock Changes in impacts of climate extremes: human systems and ecosystems.
\newblock In {\em Managing the risks of extreme events and disasters to advance climate change adaptation special report of the intergovernmental panel on climate change}, pages 231--290. Intergovernmental Panel on Climate Change, 2012.

\bibitem{EPRI2020}
Electric Power~Research Institute.
\newblock {\em Electric Power System Resiliency: Challenges and Opportunities}, 2020.

\bibitem{Jufri2019}
Fauzan~Hanif Jufri, Victor Widiputra, and Jaesung Jung.
\newblock State-of-the-art review on power grid resilience to extreme weather events: Definitions, frameworks, quantitative assessment methodologies, and enhancement strategies.
\newblock {\em Applied Energy}, 239:1049--1065, 2019.

\bibitem{mao2024valid}
Huiying Mao, Ryan Martin, and Brian~J Reich.
\newblock Valid model-free spatial prediction.
\newblock {\em Journal of the American Statistical Association}, 119(546):904--914, 2024.

\bibitem{meinshausen2006quantile}
Nicolai Meinshausen and Greg Ridgeway.
\newblock Quantile regression forests.
\newblock {\em Journal of machine learning research}, 7(6), 2006.

\bibitem{Executive2013}
Executive~Office of~the President.
\newblock {\em Economic Benefits of Increasing Electric Grid Resilience to Weather Outages}, 2013.

\bibitem{orencio2013localized}
Pedcris~M Orencio and Masahiko Fujii.
\newblock A localized disaster-resilience index to assess coastal communities based on an analytic hierarchy process (ahp).
\newblock {\em International Journal of Disaster Risk Reduction}, 3:62--75, 2013.

\bibitem{Panteli2015modeling}
Mathaios Panteli and Pierluigi Mancarella.
\newblock Modeling and evaluating the resilience of critical electrical power infrastructure to extreme weather events.
\newblock {\em IEEE Systems Journal}, 11(3):1733--1742, 2015.

\bibitem{Reinhart2018}
Alex Reinhart et~al.
\newblock A review of self-exciting spatio-temporal point processes and their applications.
\newblock {\em Statistical Science}, 33(3):299--318, 2018.

\bibitem{sklearn_quantile}
Jasper Roebroek.
\newblock Sklearn-quantile, 2022.
\newblock (visited on 2023-01-11).

\bibitem{Roege2014}
Paul~E. Roege, Zachary~A. Collier, James Mancillas, John~A. McDonagh, and Igor Linkov.
\newblock {Metrics for energy resilience}.
\newblock {\em Energy Policy}, 72(C):249--256, 2014.

\bibitem{Shapiro2017}
Leonard Shapiro, Mark Berman, Katie Zezima, and Aaron~C. Davis.
\newblock {\em Power still out at dozens of Florida nursing homes as investigation continues into 8 deaths}, 2017.

\bibitem{vovk2005algorithmic}
Vladimir Vovk, Alexander Gammerman, and Glenn Shafer.
\newblock {\em Algorithmic learning in a random world}, volume~29.
\newblock Springer, 2005.

\bibitem{xu2024conformal}
Chen Xu, Hanyang Jiang, and Yao Xie.
\newblock Conformal prediction for multi-dimensional time series by ellipsoidal sets.
\newblock {\em arXiv preprint arXiv:2403.03850}, 2024.

\bibitem{xu2021conformal}
Chen Xu and Yao Xie.
\newblock Conformal prediction interval for dynamic time-series.
\newblock In {\em International Conference on Machine Learning}, pages 11559--11569. PMLR, 2021.

\bibitem{xu2023sequential}
Chen Xu and Yao Xie.
\newblock Sequential predictive conformal inference for time series.
\newblock In {\em International Conference on Machine Learning}, pages 38707--38727. PMLR, 2023.

\bibitem{zhu2021quantifying}
Shixiang Zhu, Rui Yao, Yao Xie, Feng Qiu, Xuan Wu, et~al.
\newblock Quantifying grid resilience against extreme weather using large-scale customer power outage data.
\newblock {\em arXiv preprint arXiv:2109.09711}, 2021.

\end{thebibliography}
\end{document}